# Conditioning Optimization of Extreme Learning Machine by Multitask Beetle Antennae Swarm Algorithm

Xixian Zhang, Zhijing Yang[*], Faxian Cao, Jiangzhong Cao, Meilin Wang, Nian Cai,

*Email: yzhj@gdut.edu.cn

**Abstract**: Extreme learning machine (ELM) as a simple and rapid neural network has been shown its good performance in various areas. Different from the general single hidden layer feedforward neural network (SLFN), the input weights and biases in hidden layer of ELM are generated randomly, so that it only takes a little computation overhead to train the model. However, the strategy of selecting input weights and biases at random may result in ill-posed problem. Aiming to optimize the conditioning of ELM, we propose an effective particle swarm heuristic algorithm called Multitask Beetle Antennae Swarm Algorithm (MBAS), which is inspired by the structures of artificial bee colony (ABS) algorithm and Beetle Antennae Search (BAS) algorithm. Furthermore, the proposed MBAS is applied to optimize the input weights and biases of ELM. Experiment results show that the proposed method is capable of reducing the condition number and regression error simultaneously, and achieving good generalization performance.

1. Introduction

Extreme learning machine (ELM) proposed by G.B. Huang [1], is a feasible single hidden layer feedforward network (SLFN). It is composed of three core components: input layer, hidden layer and output layer. It has been successfully applied for many research and engineering problems. As a flexible and fast SLFN, the input weights and biases in hidden layer of ELM are assigned randomly, making the training speed increase greatly and process huge size of data in short time. There is no doubt that ELM is a good choice to cope with the tasks requiring instantaneity. Wong et al utilizes ELM to detect the real-time fault signal of gas turbine generator system [2]. Xu et al proposed a predictor based on ELM model for immediate assessment of electrical power system [3][4]. Meanwhile, it has been proved that the performances of ELM and its variants are superior to most of classical machine learning methods in the field of image processing [5][6][27], speech recognition [7][8][9], biomedical sciences [10][11][12], and so on.

There are plenty of efforts have been laid emphasis on enhancing the accuracy of ELM by means of adjusting neural network architecture and changing the numbers of hidden layer by certain rules. An incremental constructive ELM, adding its hidden nodes on the basic of convex optimization means and neural network theory, was proposed by Huang et al in 2007 [13]. While Rong et al dropped the irrelevant hidden nodes by calculating the correlation of each node with statistical approach [14]. This pruned-ELM not only raises the precision but also reduces the computation cost. It is inappropriate to evaluate the performance of a model by just considering the testing accuracy. Because the stability is also one of the most significant criterions to evaluate machine learning models. Choosing the coefficient randomly in hidden nodes is unable to ensure the numerical stability of ELM. In contrast, it will increase the risk of ill-conditioned problem, implying that the output of network may tremendously change even though the input values appear slight fluctuation.

To obtain well-conditioned ELM network, heuristic algorithms are generally adopted to optimize the parameters in hidden nodes. Recently, a promising heuristic algorithm called Beetle antennae search algorithm (BAS) was proposed [20]. The Beetle model has two antennas in different directions to detect the new solution. Beetle model always moves in the direction of antennae with the better result. Even though its mechanism of velocity alteration is uncomplicated, it still shows very good performances in many applications. Nevertheless, it is hard for BAS to find an acceptable solution in the case of non-ideal scene, since the searching ability is sensitive to the initial length of step. If the length is too large, it may miss the globally optimal solution. The small size of step will lead to "false convergence" cause of decreasing step.

Beetle in BAS always moves toward the better direction. Since improving the conditioning of ELM network is not a unimodal problem, it will be easy to fall into the local optimal solution by single beetle particle. In order to enhance the searching ability of BAS, we attempt to increase more beetle particles. Different from Particle Swarm Optimization (PSO) [16], we propose a novel particle swarm algorithm called Multitask Beetle Antennae Swarm Algorithm (MBAS) based on the frame of Artificial Bee Colony algorithm (ABC) [22], where different particles have different update rules. We add some follower particles and explore particles to enlarge the searching range and prevent falling in local optimal solutions. Some particle swarm based algorithms [17-19] have adopted to increase the accuracy of ELM. Analogously, we can put condition number of ELM into the fitness function of MBAS to optimize input weights and biases in hidden layer.

The main contributions of this paper can be summarized as follows: (1) a novel beetle swarm optimization algorithm as the extending version of BAS is proposed to enhance the ability of searching the optimal solution. A beetle group with fixed population is defined, where each beetle has different function to enlarge the searching route around the solution. (2) An improved ELM named Multitask Beetle Antennae Swarm Algorithm Extreme learning machine (MBAS-ELM) is then proposed, by combining the beetle swarm algorithm with ELM to optimize the parameters of hidden nodes. Experiment results show that MBAS-ELM is available to lower the condition number as well as testing error for regression. More details about MBAS and MBAS-ELM will be discussed in Section 3.

The remainder of this paper is scheduled as follows. In Section 2, we introduce outline of ELM and BAS. In Section3, the proposed MBAS and MBAS-ELM are introduced in details. In Section 4, results and discussion of experiments are given. Finally, conclusion of our works is shown in Section 5.

## 2. Basic Concept

### 2.1 Extreme Learning Machine

Extreme learning machine (ELM) presented by G.B. Huang, contains three main layers. Its network structure is given as Fig.1.

Suppose the size of the training data is $N \times d$, where $N$ denotes the number of samples, and $d$ denotes the dimension of features. Let $x^i = [x_1^i, x_2^i, ..., x_d^i]$ and $y^i = [y_1^i, y_2^i, ..., y_m^i]$ denote the $i^{th}$ sample and target respectively, where $m$ is the size of target vector. Then the output of the $j^{th}$ node in the hidden layer is given as follows:

$$h_j = g(W_j^i x_i^T + b_j) \tag{1}$$

where $g(\cdot)$ denotes the activation function in the hidden layer, $W_j^i = [\omega_{1,j}^i, \omega_{2,j}^i, \dots, \omega_{d,j}^i]$ is an $d$ dimensional input weight vector pointing to the $j^{th}$ hidden node, and $b_j$ is the hidden bias belonging to the $j^{th}$ hidden node. The input weights and biases are assigned randomly. The $k^{th}$ node in the output layer can be described as follows:

$$y^k = \sum_{j=1}^{L} h_j \beta_{j,k}^T \tag{2}$$

where $\beta_{j,k}^T = [\beta_{1,k}, \beta_{2,k}, \dots, \beta_{L,k}]^T$ is the output weights vector, which connects to the $k^{th}$ ouput-node, $L$ is the number of hidden nodes. The above equations can be rewritten into a matrix form as follows:

$$Y = H\beta \tag{3}$$

where $H = \begin{bmatrix} g(W_1\ x_1^T + b_1) & \cdots & g(W_L\ x_1^T + b_L) \\ \vdots & \ddots & \vdots \\ g(W_1\ x_N^T + b_1) & \cdots & g(W_L\ x_N^T + b_L) \end{bmatrix}_{N \times L}$, $\beta = \begin{bmatrix} \beta^1 \\ \vdots \\ \beta^L \end{bmatrix}_{L \times m}$, $Y = \begin{bmatrix} y^1 \\ \vdots \\ y^m \end{bmatrix}_{N \times m}$

Finally, the output weight $\beta$ can be calculated analytically by

$$\beta = H^\dagger Y \tag{4}$$

where $H^\dagger$ denotes the Moore-Penrose (MP) inverse of the hidden layer matrix $H$. It should be rewritten when matrix $H$ is nonsingular as

$$\beta = (H^T H)^{-1} H^T Y \tag{5}$$

The essence of training ELM network is to estimate the output weight $\beta$ with analytically method.

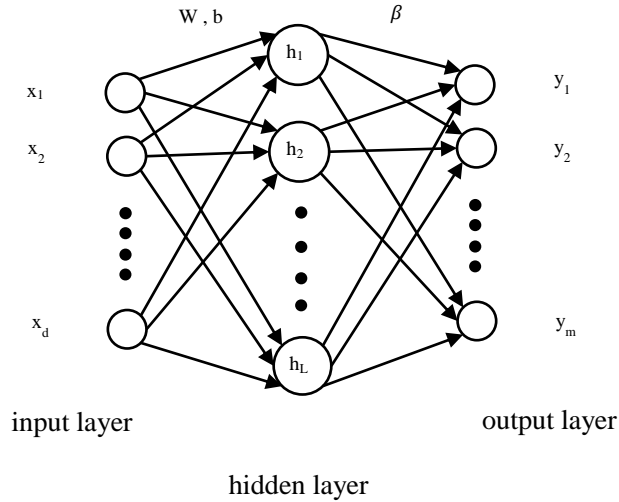

**Fig. 1** Architecture of ELM network

**2.2 Beetle Antennae Search Algorithm**

BAS algorithm is the simulation of beetle foraging. Suppose that a beetle is in the $k$ dimension space. And its position denotes as $P_{centre}(p_1, p_2, ..., p_k)$, where $P_{centre} \epsilon \mathcal{R}^k$. Two antennae endpoints of beetle are on right and left sides, denotes as $P_{right}$, $P_{left}$, where $P_{right} \epsilon \mathcal{R}^k$, $P_{left} \epsilon \mathcal{R}^k$. $P_{centre}$, $P_{right}$, $P_{left}$ are on the same line. Antennas are used to receive odors signal from environment. Beetle will move to next position where smell signal is more intense. For example, odors signal on right side is more intense, beetle renews position in the direction of $P_{right}$.

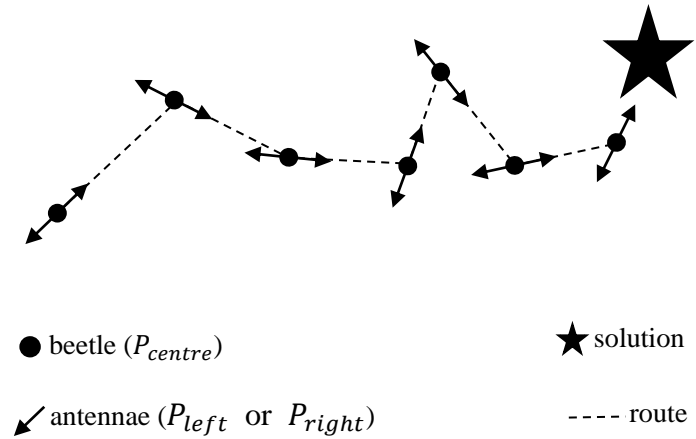

● beetle ($P_{centre}$)  ★ solution

↗ antennae ($P_{left}$ or $P_{right}$)   ---- route

**Fig.2** graphic description of BAS searching process

When the $t^{th}$ iteration begins, antennas direction $Dir^t$ is assigned at random. We define an unit vector $\overrightarrow{P_{centre}P_{right}}$ as the positive direction, then $Dir^t$ can be defined as equation (6).

$$Dir^t = \frac{rand(k)}{|rand(k)|} = \overrightarrow{P_{centre}P_{right}} = -\overrightarrow{P_{centre}P_{left}} \qquad (6)$$

where $rand(\cdot)$ represents the random function to generate a $k$ dimensions vector, $|\cdot|$ denotes the modulo operation.

The endpoint positions in the $t^{th}$ iteration are calculate by the following equations

$$P_{right}^t = P_{centre}^t + d_a * Dir^t \qquad (7)$$

$$P_{left}^t = P_{centre}^t - d_a * Dir^t \qquad (8)$$

where $d_a$ represents the distance among beetle and its antennae endpoints.

In the case of searching minimum, beetle updates its position as

$$P_{centre}^{t+1} = P_{centre}^t - \delta^t * Dir^t * sign(f(P_{right}^t) - f(P_{left}^t)) \qquad (9)$$

where $f(\cdot)$ is the fitness function of BAS algorithm, $sign(\cdot)$ is the sign function, $\delta^t$ denotes length of step in the $t^{th}$ iteration.

Finally, update the length of step and antennae according to the following two equations

$$d_a^{t+1} = \eta * d_a^t \;,\; if \; d_a^t > d_a^{min} \qquad (10)$$

$$\delta^{t+1} = \eta * \delta^t \;,\; if \; \delta^t > \delta_{min} \qquad (11)$$

where $\eta$ denotes the decay factor, $0 < \eta < 1$, $d_a^{min}$ and $\delta_{min}$ respectively denote the minimum size of antennae and step.

The BAS algorithm for searching minimum is summarize in Algorithm 1.

| Algorithm 1: BAS algorithm for searching minimum | |
| --- | --- |
| Initialization | Input $\eta$, $d_a^{min}$, $\delta_{min}$, $d_a^0$, $\delta^0$, $t = 0$, $T_{max}$ <br> Randomly generate $P_{centre}^0(p_1, p_2, \dots, p_k)$ |
| Iteration | While ( $t < T_{max}$ ) or (stop criterion) <br> $Dir^t = \dfrac{rand(k)}{|rand(k)|}$ <br> $P_{right}^t = P_{centre}^t + d_a * Dir^t$ <br> $P_{left}^t = P_{centre}^t - d_a * Dir^t$ <br> $P_{centre}^{t+1} = P_{centre}^t - \delta^t * Dir^t * sign(f(P_{right}^t) - f(P_{left}^t))$ <br> $d_a^{t+1} = \eta * d_a^t$ , if $d_a^t > d_a^{min}$ <br> $\delta^{t+1} = \eta * \delta^t$ , if $\delta^t > \delta_{min}$ <br> If $f(P_{centre}^{t+1}) < f(P_{centre}^t)$ <br> $\quad P_{best} = P_{centre}^{t+1}$ <br> $\quad F_{best} = f(P_{centre}^{t+1})$ <br> End if <br> t = t+1 <br> End While |
| Return | $P_{best}$ & $F_{best}$ |

## 3. The Proposed Method

### 3.1 The Proposed Beetle Swarm Optimization Algorithm

The seeking capability of single beetle particle is limited, especially for complex tasks since BAS tends to sink into local optimal solution or incapable of obtaining adequate solutions. The idea of particle swarm optimization increases the number of searching in each iteration, hence there is higher possibility to find better solution. T.T. Wang [21] presented a type of beetle swarm algorithm for multi-objective optimization which had shown its competitive performance. It was shown that the hybrid of BAS and swarm optimization is worth of study.

In human society, everyone has his own specific vocation, working in cooperation with each other. In order to complete a complex mission, people on different occupation make full use of their respective advantages to maximize the efficiency. Correspondingly, the division of labor widely exists in animal kingdom, especially the gregarious animals. There are several works follow the rule of division of labor, such as Artificial Bee Colony algorithm (ABC) and Chicken Swarm Optimization algorithm (CSO) [23]. In general, this type of swarm algorithms creates a population with fixed number of particle. Each particle in this group is appointed to play specified role, updating their position with different rule, which enriches the diversity of particle moving methods. The above process simulates how individual exerts its function and interdependent collaboration.

Inspired by ABC algorithm, we propose an individual cooperation based beetle swarm optimization algorithm namely Multitask Beetle Antennae Swarm Algorithm (MBAS). Details are given as follows.

Firstly, we define a beetle particle swarm with the population number $N$. The main vocations of beetle particle in this group are searchers, follower and explorer, respectively. For searchers, the most important mission is to search the solution in the feasible set and renew the position with BAS algorithm. The alteration of the position is given by the following equation:

$$P_S^{t+1}(\omega, b) = BAS\_move(P_S^t(\omega, b), L_S^t, L_S^{min}, d_a^0, d_a^{min}, \eta) \tag{12}$$

where $BAS\_move(\cdot)$ represents the moving approach in the BAS algorithm in algorithm 1. For convenience, let $d_a^{min} = 0$.

For followers, they pursue certain searcher beetle step by step to find the latent optimal solutions around the current global optimal solution. For convenience, all the follower particles follow searcher particle with the best fitness values in our work. The corresponding movements are as follows:

$$D_F^t = P_S^t(\omega, b) - P_F^t(\omega, b) \tag{13}$$

$$P_F^{t+1}(\omega, b) = P_F^t + \frac{D_F^t}{|D_F^t|} * L_F \tag{14}$$

where $L_F$ is the length of the step of the follower, $P_S^t(\omega, b)$ is the position of the best searcher particle.

For explorers, they move at random with fixed step length, preventing the whole system to fall into local optimal solution. The equations are as follows:

$$D_E^t = \frac{rand(\dim(D_E^t))}{|rand(\dim(D_E^t))|} \tag{15}$$

$$P_E^{t+1} = P_E^t + D_E^t * L_E \tag{16}$$

where $L_E$ denotes the step size of explores.

Secondly, determine the numbers of each beetle with three vocations respectively as $N_S$, $N_F$ and $N_E$ with $N_S + N_F + N_E = N$. To generate the position $P_i(\omega, b)$ of each beetle particle randomly, $i = 1,2,...,N$, calculate the fitness value $F_i(P_i(\omega, b))$ of each particle, sort $F_i(P_i(\omega, b))$ from smallest to largest. Then appoint the top $N_S$ particle as searchers, the top $N_S + 1$ to $N_S + N_F$ as followers. The remaining particles are explorers.

Thirdly, the follower beetles approach the searcher beetles with the best fitness value. All the searcher beetles and explorer beetles move to the next position in their own ways. Calculate new fitness values with new positions for all particles. Redistribute the vocation to every beetle according to the new fitness values at the present iteration. If the best fitness value in current iteration is better than the previous best one, the current best fitness value replaces the best value.

Finally, the above process is repeated until $N_I$ epochs are completed. Return the best position and best fitness value. The MBSA algorithm is summarized in Algorithm 2.

| Algorithm 2: MBSA algorithm | |
| --- | --- |
| **Step 1:** | Initialize all the Parameter |
| **Step 2:** | Generate the position $P_i(\omega, b)$, where $i = 1,2,...,N$ |
| **Step 3:** | Calculate the fitness value $F_i(P_i(\omega, b))$, where $i = 1,2,...,N$ |

| Step 4: | Sort $F_i(P_i(\omega, b))$ from smallest to largest ,then obtain $F_j(P_j(\omega, b))$ where j = 1,2,…,N |
|---|---|
| Step 5: | Form j = 1 to $N_S$, appoint $P_i(\omega, b)$ as $P_S$ |
| | Form j = $N_S$ +1 to $N_S + N_F$, appoint $P_i(\omega, b)$ as $P_F$ |
| | Form j = $N_S + N_F$ +1 to $N_S + N_F + N_E$, appoint $P_i(\omega, b)$ as $P_E$ |
| Step 6: | For t =1 to $N_I$ |
| |     For m = 1 to $N_F$ |
| |         Direction update as equation (13) |
| |         Position update as equation (14) |
| |     End for |
| |     For m = 1 to $N_S$ |
| |         Direction and position update as equation (12) |
| |     End for |
| |     For m = 1 to $N_E$ |
| |         Direction randomly chosen as equation (15) |
| |         Position update as equation (16) |
| |     End for |
| |     Repeat **Step 3** to calculate the fitness values $F_i^t(P_i^t(\omega, b))$, where i = 1,2,…,$N_S + N_F + N_E$ |
| |     If $\min(F_i^t) < F_{best}$ |
| |         $F_{best} = \min(F_i^t)$ |
| |         $P_{best} = P_i^t$ |
| |     End If |
| |     Repeat **Step 4** & **Step 5** to reassign vocation for all particles according to latest fitness values. |
| | End for |
| Step 7: | Return $F_{best}$ & $P_{best}$ |

In this section, we introduce the details of MBAS. Different from reference [20], we do not use the frame of PSO to modify BAS. In the frame of PSO, the update strategy of all particles is the same. If all particles gather into a small space where the global solution is not inside, then no particle can jump out from this space. However, in ABC and MBAS, some particles move randomly to prevent the particles gather together. Compared with ABC algorithm, MBAS uses BAS to simplify the foraging process of employed bees. We add several particles with different functions to improve the searching ability of BAS.

**3.2 Beetle Swarm Optimization Extreme Learning Machine**

The original ELM is with no need for training hidden layer biases and input weights. Although this mechanism can greatly reduce the computation time, it restricts the stability and performance. The optimization of parameters in hidden layer of ELM is a complex problem. The searching ranges BAS is limited by particle number. In MBAS, follower particles and explore particles are applied to assist the searcher particles to find more potential solution around the best solution. Position interaction between different particles become stronger, thus MBAS is suitable for complex optimization problems.

In this section, we introduce a new type ELM namely Multitask Beetle Antennae Swarm

Algorithm Extreme learning machine (MBAS-ELM), and the flowchart of training MBAS-ELM is given in Fig. 3. The key of MBAS-ELM is choosing a suitable fitness function to satisfy the criterion. In order to build a reliable network, four criteria (denoted as $RMSE$, $R^2$, $K_2(H)$, $\|\beta\|_2$) are considered as our optimal targets.

Root mean squared error (RMSE) and coefficient of determination ($R^2$) are two most classical criterions for regression. $RMSE$ is used to measure distance among predicted value and ground-truth. The smaller the RMSE value is, the better the performance is. RMSE is defined as follows.

$$RMSE = \sqrt{\frac{\sum_{i=1}^{N}(\hat{y}_i - y_i)^2}{N}} \tag{17}$$

where $\hat{y}_i$ is the $i^{th}$ prediction in ELM, $y_i$ is the $i^{th}$ ground-truth, and $N$ is the total number of prediction.

$R^2$ is applied to measure fitting degree between predicted value and ground-truth. The closer the $R^2$ is to 1, the better the fitting degree is.

$$R^2 = 1 - \frac{\sum_{i=1}^{n}(\hat{y}_i - y_i)^2}{\sum_{i=1}^{n}(y_i - \bar{y})^2} \tag{18}$$

where $\hat{y}_i$ is the $i^{th}$ prediction, $y_i$ is the $i^{th}$ ground truth, and $\bar{y}$ is the mean of all ground truth.

Furthermore, the norm of output weight $\|\beta\|_2$ and condition number $K_2(H)$ are also the significance criterion for ELM regression. The generalization ability is closely connected to the norm of output weight $\|\beta\|_2$ [24][25]. The smaller the norm value is, the better the generalization performance is. G.P. Zhao [26] employs the condition number $K_2(H)$ to measure the stability of ELM. The global optimum solution of condition number is 1. A matrix with high condition number implys that it's ill-conditioned. Its definition is given as follows

$$K_2(H) = \sqrt{\frac{\lambda_{max}(H^T H)}{\lambda_{min}(H^T H)}} \tag{19}$$

where $\lambda_{max}(H^T H)$ and $\lambda_{min}(H^T H)$ are the largest and the smallest eigenvalue of matrix $H^T H$, respectively. The fitness function $\mathcal{F}()$ is constituted of above four criterions, which is given by the following equation,

$$\mathcal{F}(\ ) = \|\beta\|_2 * K_2(H) + \gamma * (1 - R^2) \tag{20}$$

where $\gamma$ is a scaling factor. The reason why we design such fitness function will be explained in Section 4.3.

The training procedure of MBAS-ELM can be summarized as follows:

**Step1**: Given training data $X_{train}$, training label $Y_{train}$, testing data $X_{test}$, testing label $Y_{test}$. We divide $X_{train}$ and $Y_{train}$ into three non-intersect subset $(X_v^r, Y_v^r)$ for 3-fold cross-validation, where r = 1,2,3.

**Step2**: Fix the size of population. Randomly initialize position of each particle position, which made up of input weights and biases in the $r^{th}$ subset, described as,

$$P_i\ (\omega, b) = [\omega_{11}, \omega_{12}, \dots, \omega_{1L}, \omega_{21}, \omega_{22}, \dots, \omega_{2L}, \omega_{n1}, \omega_{n2}, \dots, \omega_{nL}, b_1, b_2, \dots, b_L]_{(n+1) \times L} \tag{21}$$

where $n$ is the size of input layer, $L$ denotes the number of hidden layer, $P_i\ (\omega, b) \in [-1, 1]$,

**Step3**: Calculate fitness function and RMSE for all particle in each iteration. Select the position corresponding to the smallest fitness value in current iteration as $P_{current}$. The best positon $P_{best}$ renew as following equation,

$$P_{best} = \begin{cases} P_{current}, & \mathcal{F}(P_{current}) < \mathcal{F}(P_{best}) \text{ and } RMSE_{current} < RMSE_{best} \\ P_{best}, & else \end{cases} \quad (22)$$

**Step4**: Repeat Step3 for each subset, select the best position $P_{best}^r$, $r = 1,2,3$.

**Step5**: Employ $\hat{P}_{best}(\hat{\omega}, \hat{b})$ obtained from Step 4 to estimate the trained network.

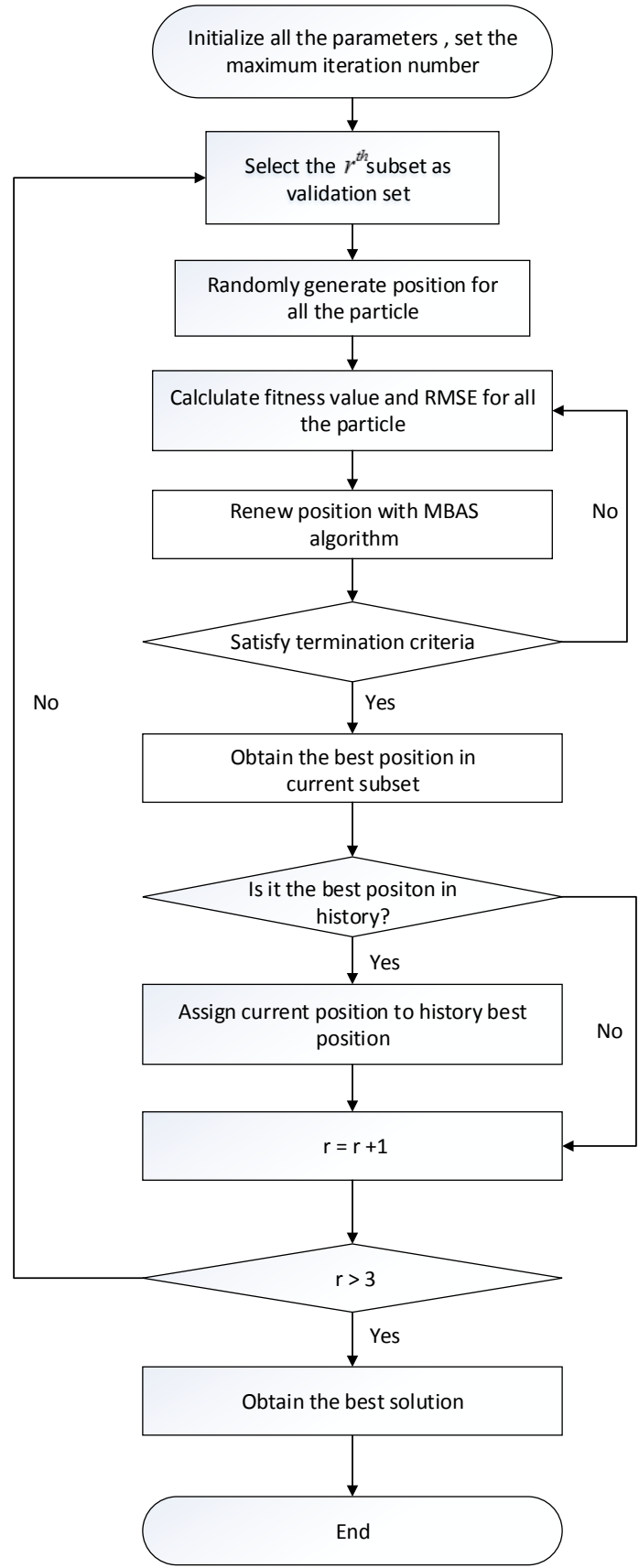

**Fig.3**. flowchart of MBAS-ELM training process

## 4. Experiment Verification

### 4.1 Experiment Setup

In this section, to test the performances of the proposed MBAS-ELM, we use ELM and BAS-ELM algorithms as the comparisons on two real-world regression benchmark datasets. Details of the datasets are given in Table 1. All the datasets are available from LIBSVM dataset [28] and UCI dataset [29]. Four criterions ($RMSE$, $R^2$, $K_2(H)$, $\|\beta\|_2$) discussed in Section 3.2 are adopted to estimate the performances of all algorithms. The influence of hidden node number is not the emphases in this paper, thus we empirically assign 10 nodes for each ELM network. BAS-ELM can be seen as a particular case of MBAS-ELM, which only contains one searcher particular.

**Table** 1 Information of real-world regression dataset

| Names | Features | Training Data | Test Data |
|---|---|---|---|
| bodyfat | 14 | 100 | 152 |
| housing | 13 | 250 | 256 |

### 4.2 Experiment Results

Three aforementioned ELM algorithms (original ELM, BAS-ELM, MBAS-ELM) are independently estimated 30 times on two datasets severally with the same training and test data in each epoch. The training data and testing data are chosen randomly at the beginning of each epoch. The outputs of three algorithms are normalized to the range from 0 to 1. After 30 times estimations, we calculate the mean of $RMSE$, $R^2$, $K_2(H)$, $\|\beta\|_2$, and the standard deviation of $RMSE$. For the parameter $\gamma$ in Eq. (20), we use the cross validation to select $\gamma$ with the smallest fitness value and RMSE. The experiment results are given in the following tables and figures.

**Table** 2   Performances of three ELM algorithms on bodyfat dataset (Best result is marked in bold type)

| Names | RMSE | standard deviation | coefficient of determination $R^2$ | Condition Number $K_2(H)$ | Norm $\|\beta\|_2$ |
|---|---|---|---|---|---|
| ELM | 0.0723 | 0.0184 | 0.8025 | 110.8503 | 2.0563 |
| BAS-ELM | 0.0512 | 0.0078 | 0.9030 | 61.9109 | 1.1241 |
| MBAS-ELM | **0.0431** | **0.0073** | **0.9310** | **51.0597** | **1.0399** |

**Table** 3   Performances of three ELM algorithms on housing dataset (Best result is marked in bold type)

| Names | RMSE | standard deviation | coefficient of determination $R^2$ | Condition Number $K_2(H)$ | Norm $\|\beta\|_2$ |
|---|---|---|---|---|---|
| ELM | 0.1361 | 0.0168 | 0.5559 | 61.8257 | 1.4522 |
| BAS-ELM | 0.1291 | 0.0102 | 0.6016 | 35.1917 | 1.1343 |
| MBAS-ELM | **0.1208** | **0.0101** | **0.6522** | **28.8559** | **0.9800** |

As can be seen in table 2 and table 3, Two optimized ELM algorithms (BAS-ELM, MBAS-ELM) can achieves promotion in *RMES*, $R^2$, $K_2(H)$ and $\|\beta\|_2$. The proposed algorithm acquire the best performance compared with other algorithms, especially on bodyfat dataset.

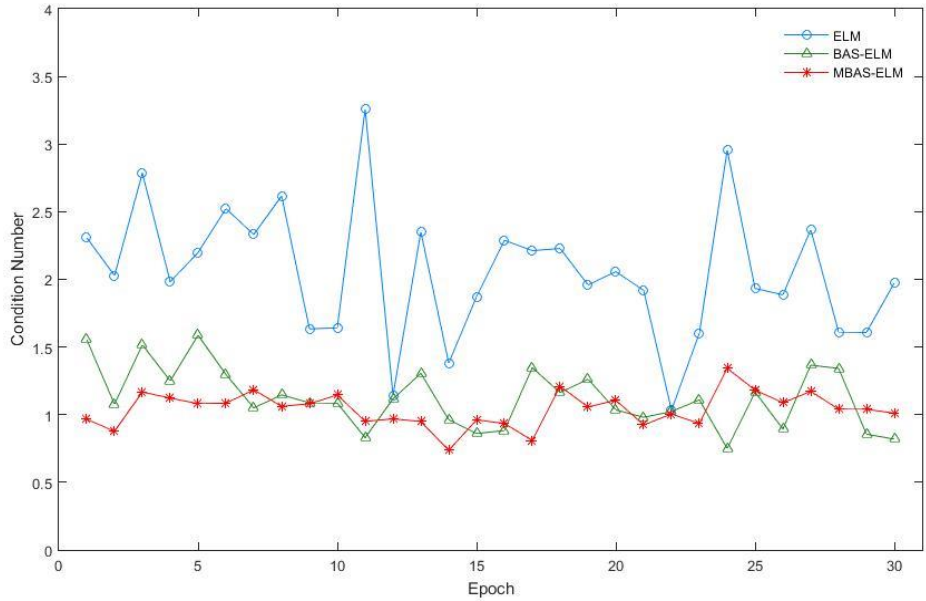

**Fig.4**. norm value comparison of three algorithms on bodyfat dataset

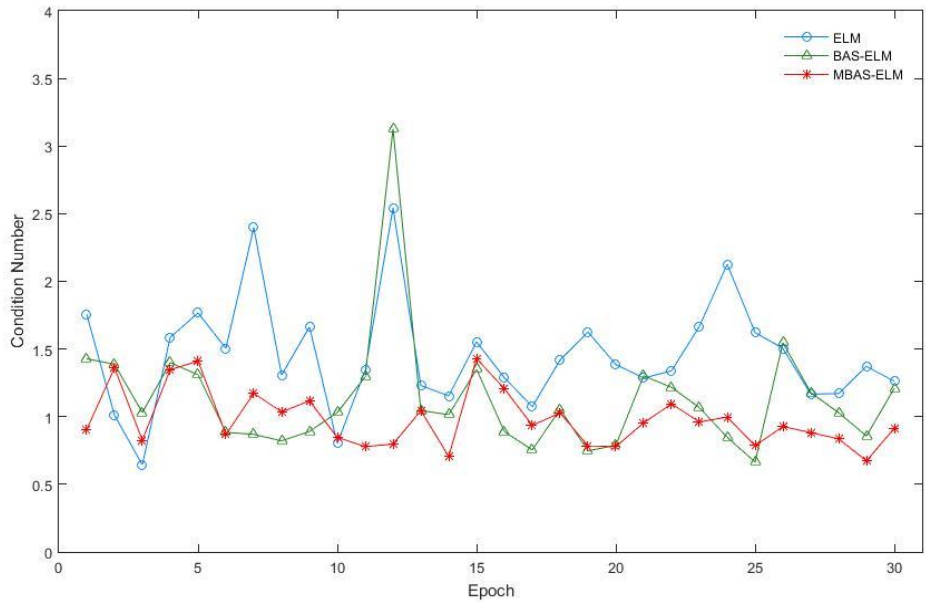

**Fig.5**. norm value comparison of three algorithms on housing dataset

The norm values of the three methods in the two datasets are given in Figs. 4-5, respectively. The variation curve of the original ELM appears great fluctuation. Two optimized ELM algorithms (BAS-ELM, MBAS-ELM) can reduce the fluctuation and the mean of norm value with different degrees. Moreover, the proposed MBAS-ELM achieves the smallest mean on all datasets.

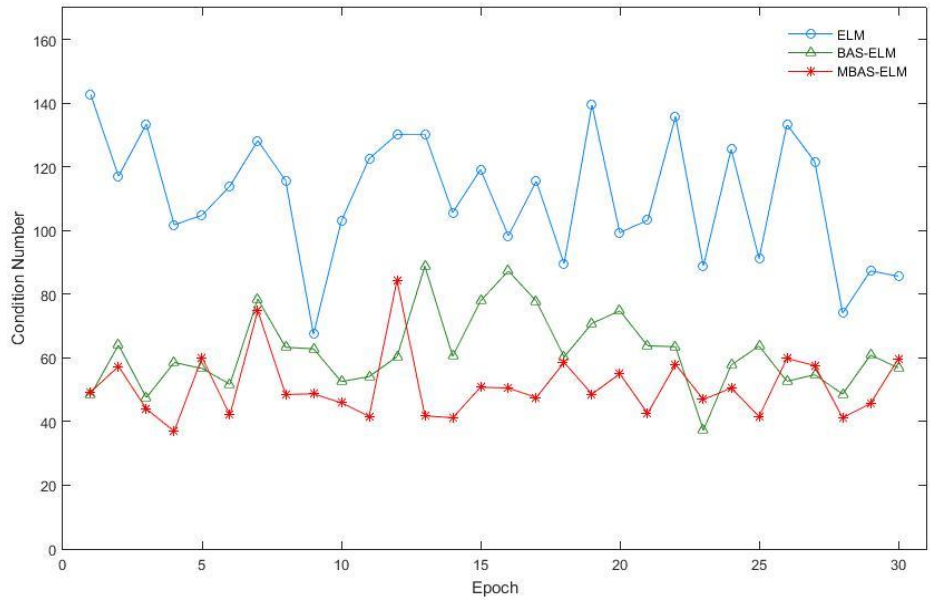

**Fig.6**. condition number comparison of three algorithms on bodyfat dataset

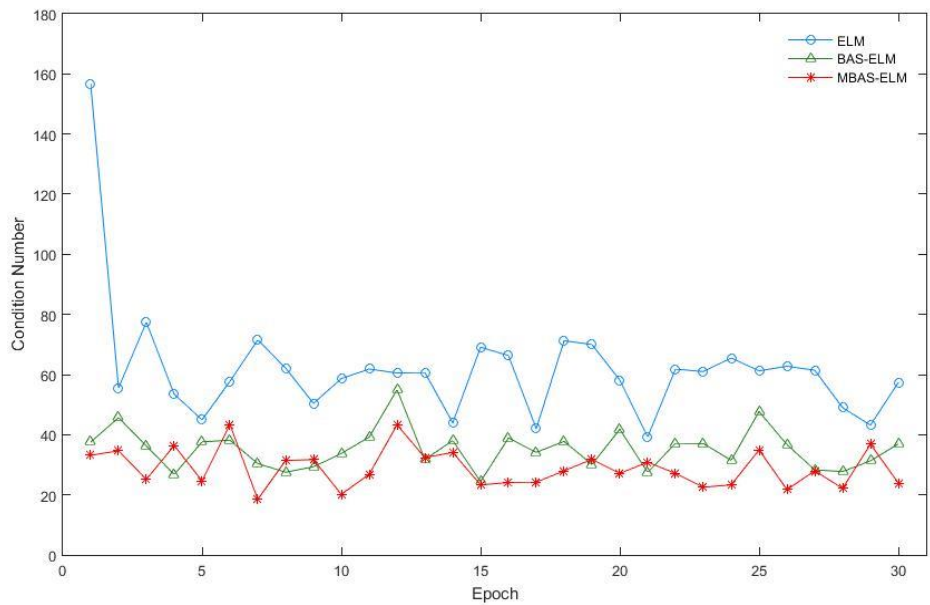

**Fig.7**. condition number comparison of three algorithms on housing dataset

The condition numbers of the three methods in the two datasets are given in Figs. 6-7, respectively. Not surprisingly, the original ELM appears the maximum fluctuation on all datasets, due to the random parameter in hidden layer. Two optimized ELM algorithm (BAS-ELM, MBAS-ELM) can reduce the risk of ill-conditioned in different degrees. The proposed MBAS-ELM can achieve the best performance in system stability.

## 4.3 Experiment Discussion

In order to fully improve the conditioning and generalization ability, we consider the $K_2(H)$ and $\|\beta\|_2$ as variables in our fitness function as shown in equation (20). Meanwhile, we add $R^2$ to ensure the regression precision. From the definition equation of RMSE and $R^2$, we can easily find that they contain the same component $\sum_{i=1}^{n}(\hat{y}_i - y)^2$, so we can tune $R^2$ instead of RMSE to lower the error. If a few of predicting points with huge error appear in regression result, the mean error may not give rise to great changes, but the fitting degree may probably decrease. For regression task, we hope the fitting degree can be as close to 1 as possible. Moreover, the scaling factor $\gamma$ of equation (20) can be seen as a weight of accuracy and conditioning. If $\gamma$ is too small, the MBAS algorithm tends to search the solution with very low condition number and norm value. On the contrary, large $\gamma$ probably leads to ignore the condition number and norm value. In general, the range of RMSE is uncertain. Though we normalize the output from 0 to 1, the range of RMSE is still uncertain if the prediction number is too large. It's hard to determine $\gamma$ when RMSE is in a large range. Different form RMSE, the range of norm value is usually in certain range, so we can easily determine $\gamma$ to change the proportion of precision and stability according to actual needs. Hence, $R^2$ should be considered in the fitness function.

From above experiment results, the proposed ELM algorithm shows its great performance in most datasets. MBAS-ELM can ensure the regression precision within acceptable range and greatly enhance stability and generalization ability.

## 5. Conclusion

In this paper, a swarm optimization algorithm called MBAS algorithm is proposed. Inspired by the existing swarm optimization algorithm, the particles play different roles in MBAS to enlarge the searching area. Furthermore, we apply MBAS algorithm to optimize the input weights and biases in hidden layer of ELM to improve conditioning and generalization ability. Experiments are established to show the performance of the proposed MBAS-ELM algorithm by comparing with the original ELM and BAS-ELM. Experiment results show that MBAS-ELM can achieve the best performance on all the datasets, and effectively improve the conditioning and generalization ability. Since MBAS-ELM contains more beetle particles than BAS-ELM, it needs relatively more time for computation. Fortunately, BAS is a fast and simple heuristic algorithm. Therefore, the computation cost can stay in an acceptable level by choosing reasonable particle population at the beginning of MBAS-ELM.

**Acknowledgement**

This work is supported in part by the National Nature Science Foundation of China (no. 61471132), the Training program for outstanding young teachers of Guangdong Province (no. YQ2015057), Science and Technology Planning Project of Guangdong Province (no. 2017B090901056), Science and Technology Program of Guangzhou, China (nos. 201803010065, 201802020010).